# Unsupervised deep learning for grading age-related macular degeneration using retinal fundus images

Baladitya Yellapragada[1,2,3], Sascha Hornhauer[2], Kiersten Snyder[3], Stella Yu[1.2], Glenn Yiu[3]

[1]Department of Vision Science, University of California, Berkeley, Berkeley, CA
[2] International Computer Science Institute, Berkeley, CA
[3] Department of Ophthalmology & Vision Science, University of California, Davis, Sacramento, CA

Corresponding Author:
Glenn Yiu
4860 Y St., Suite 2400
Sacramento, CA 95817
gyiu@ucdavis.edu

## Abstract

Many diseases are classified based on human-defined rubrics that are prone to bias. Supervised neural networks can automate the grading of retinal fundus images, but require labor-intensive annotations and are restricted to the specific trained task. Here, we employed an unsupervised network with Non-Parametric Instance Discrimination (NPID) to grade age-related macular degeneration (AMD) severity using fundus photographs from the Age-Related Eye Disease Study (AREDS). Our unsupervised algorithm demonstrated versatility across different AMD classification schemes without retraining, and achieved unbalanced accuracies comparable to supervised networks and human ophthalmologists in classifying advanced or referable AMD, or on the 4-step AMD severity scale. Exploring the network's behavior revealed disease-related fundus features that drove predictions and unveiled the susceptibility of more granular human-defined AMD severity schemes to misclassification by both ophthalmologists and neural networks. Importantly, unsupervised learning enabled unbiased, data-driven discovery of AMD features such as geographic atrophy, as well as other ocular phenotypes of the choroid, vitreous, and lens, such as visually-impairing cataracts, that were not pre-defined by human labels.

# Introduction

Deep convolutional neural networks (CNNs) can be trained to perform visual tasks by learning patterns across hierarchically complex scales of representations [1] with earlier filters identifying low-level concepts such as color, edges, and curves, and later layers focused on higher-level features such as animals or animal parts [2]. Although CNNs are typically used for natural image tasks such as animal classification [3], aerial-view vehicle detection [4], and self-driving [5], these algorithms have also been adapted for medical image classification for clinical applications. In ophthalmology, deep learning algorithms can provide automated expert-level diagnostic tasks such as detection of diabetic retinopathy [6,7,8,9,10,11], age-related macular degeneration (AMD) [11,12,13], and glaucoma [15,16,17] using retinal fundus images. They can also extract information including age, sex, cardiovascular risk [18], and refractive error [19] that are not discernable by human experts.

However, supervised learning approaches are trained using expert-defined labels which classify disease type or severity into discrete classes based on human-derived rubrics that are prone to bias and may not accurately reflect the underlying disease pathophysiology. Because supervised networks can only identify phenotypes that are defined by human experts, they are also limited to identifying known image biomarkers. Moreover, training labels are labor intensive to generate, typically involving multiple expert graders who are susceptible to human error. Even trained ophthalmologists do not grade retinal images consistently, with significant variability in sensitivity for detecting retinal diseases [20].

Unsupervised learning organizes images based on features that are not predetermined by human graders. These algorithms enable image classification without the constraints or arbitrary delineation of human labels during training. Unsupervised and semi-supervised neural networks have been developed using several methods, including instance-based learning [21], exemplar learning [22], deep clustering [23], and contrastive learning [24,25,26,27]. An unsupervised network approach called Non-Parametric Instance Discrimination (NPID) was previously designed for complex visual tasks [24]. NPID predicts a query image's class label by determining the most common label among its nearest neighbors within a multi-dimensional hypersphere of encoded feature vectors drawn from training images. This technique significantly outperforms other unsupervised networks for ImageNet, Places, and PASCAL Visual Object Classes classification tasks [24].

In this study, we trained the NPID algorithm using retinal fundus photographs from the Age-Related Eye Diseases Study (AREDS) to classify AMD severity, then investigated the network's behavior to explore human label biases that may not conform to disease pathophysiology, and to enable unbiased discovery of ocular phenotypes. This unsupervised CNN achieved similar accuracy to ophthalmologist graders [20], and matched or exceeded the performance of published supervised CNNs, despite never learning the class definitions directly. Importantly, our examination of NPID behavior provides new insights into the visual features that drive test prediction, and enabled unbiased, data-driven discovery of AMD phenotypes not encompassed by human-assigned categories, as well as non-AMD features including camera artifacts, lens opacity, vitreous anomalies, and choroidal patterns. Our results show that unsupervised deep learning based on visual similarities rather than human-defined labels can bypass human bias and imprecision, enable accurate grading of disease severity comparable to supervised neural networks or human experts, and discover novel pathologic or physiologic phenotypes that the algorithm was not specifically trained to detect.

## Results

### Study data characteristics & partitioning

Sponsored by the National Eye Institute, the AREDS enrolled 4757 subjects aged 55 to 80 years in a prospective, randomized, placebo-controlled clinical trial to evaluate oral antioxidants as treatment for AMD. The median age of participants was 68, 56% were women, and 96% were Caucasian [28,29]. Color fundus images from AREDS were previously graded by the University of Wisconsin fundus photograph reading center for anatomic features, including the size, area, and type of drusen, area of pigmentary abnormalities, area of geographic atrophy (GA), and presence of choroidal neovascularization (CNV) [28]. These gradings were used to develop a 9+3-step AMD severity scale for each eye which predicts the 5-year progression risk to CNV or central GA [30], with steps 1-3 representing no AMD, 4-6 representing early AMD, 7-9 representing intermediate AMD, and 10-12 representing advanced AMD including central GA (step 10), CNV (step 11), or both (step 12) [31,28,29,30] (Supplemental Figure 1a). Both the 9+3-step scale and the simplified 4-step scale have been used to successfully train supervised CNNs to classify AREDS fundus images for AMD severity [20,13]. As NPID's feature space is more dependent on low-level visual variety to make its prediction space less susceptible to bias, performance is bolstered by not excluding any images, such as stereoscopic duplicates or repeated subject eyes from different visits. A total of 100,848 fundus images were available, with a long-tailed imbalance and overrepresentation of the no-AMD classes for both scales, and class 11 (CNV) in the 9+3-step scale (Supplemental Figure 1b-1c). Images were randomly partitioned into training, validation, and testing datasets in a 70:15:15 ratio, respectively, while ensuring that stereo pairs from the same eye did not appear across different datasets.

Accuracy in grading AMD severity

We first evaluated NPID performance on a 2-step discrimination task for detecting advanced AMD (CNV and/or central GA), and found that our unsupervised network achieved an unbalanced accuracy (94%) that is comparable to the performance of a published supervised CNN (96.7%) or trained ophthalmologist (97.3%) [14]. The balanced accuracy, which is more applicable due to dataset imbalance, was also similar between the unsupervised NPID (82%), supervised network (81%), and ophthalmologist (89%) (Figure 2a). Next, we compared the balanced accuracy of NPID with another supervised algorithm to distinguish "referable" AMD (intermediate or advanced) from no or early AMD, and found that our unsupervised network performed only slightly worse (87%) than the supervised network (92%) and ophthalmologist (96%) [20], despite never learning the class definitions directly (Figure 2b). For grading AMD severity using the 4-step scale, NPID achieved a 65% balanced accuracy, which was comparable to a published supervised network (63%) and ophthalmologist (67%)(Figure 2c) [20]. In particular, the confusion matrix for NPID demonstrated superior performance for distinguishing early AMD (class 2) as compared to both the supervised network and human expert (Figure 2d) [20].

When applied to a finer classification task, NPID only achieved a balanced accuracy of 25% on the 9+3-step scale, as compared with 74% using a supervised network [13] that utilized the same backbone network as our NPID approach. Among the 50 nearest neighbors in our network, 28% shared the query image's label while 68% were within 2 steps of the correct 9+3-step label (Figure 2e). Even for cases with incorrect 9+3-step class predictions, the 50 nearest neighbor images shared the query's 4-step class label 56% of the time, which accounts for the higher accuracy of our network in the 4-step classification task. Thus, although unsupervised learning achieves lower performance on the finer 9+3-step AMD severity scale compared to binary or 4-step AMD classifications, incorrect predictions deviate minimally from ground-truth labels.

Network behavior for grading AMD severity

To discern how the NPID network visually organizes images from different AMD classes, we employed t-Distributed Stochastic Neighbor Embedding (t-SNE) visualization, which maps the encoded 64-dimensional features onto 2-D coordinates. On the 4-step AMD severity scale, fundus images with no (blue), intermediate (yellow), and advanced (red) AMD formed distinct clusters, while early AMD (aqua / green) images are scattered throughout the plot (Figure 3a), which likely explains the lower performance in this class (Figure 2d). On the 9-step AMD severity scale (Figure 3b), the t-SNE plot appear similar to that of the 4-step scale, as each of the 4 major classes on the simplified scale are dominated by one or two of the finer classes within each subset (Supplemental Figure 1b), and may account for the poorer performance of our unsupervised network on the 9+3-step task.

Examining the training images that contribute to NPID predictions helps explain the unsupervised network's behavior in an interpretable way that supervised networks cannot, as the specific training images that drive a supervised network's predictions cannot be easily recovered. In our study, comparison of query images with a selection of neighboring reference images demonstrates high phenotypic similarity across adjacent 9+3-step classes (Figure 3c), and explains class confusions that contribute to NPID performance loss on the finer-grained 9+3-step scale. Furthermore, training the unsupervised network on image subsets within each of the 4-step simplified AMD classes – a process known as hierarchical learning – showed that the early AMD subset (class 2 on the 4-step scale) exhibited the least fine-class separability across the 9+3-step scale, with many class 4 images that resembled no AMD and class 6 images that appear similar to intermediate AMD (Figure 3c, bottom rows), which helps explain the difficulty with distinguishing early AMD images by the NPID method, as well as by supervised networks and human ophthalmologists (Figures 2d-2e).

To determine which AMD features contributed most to the unsupervised learning, we mapped reading center-designated labels including (1) drusen size, area, and type, (2) depigmentation or hyperpigmentation area, and (3) total or central GA area onto the tSNE plots (Figure 4). We found that drusen area provided the most visually distinct clusters that matched the separation of the 4-step severity scale. GA area and depigmentation correlated well with advanced AMD classes as expected, while larger drusen size or soft drusen type corresponded to intermediate AMD classes. Our results show that t-SNE visualizations, similarity searches, and hierarchical learning based on NPID can unveil the susceptibility of more granular human-defined AMD severity schemes to misclassification by both ophthalmologists and neural networks, and provide insight into the anatomic features that may drive AMD severity predictions.

Data-driven AMD phenotype discovery

Current AMD severity scales suffer from human bias because they were developed in part to reflect clinical severity (i.e. impact on visual function) rather than disease pathophysiology. For example, only vision-threatening GA involving the central macula was ascribed as advanced AMD (class 10 on the 9+3-step scale), while non-central GA cases were scattered across other AMD classes. With the goal of extracting features of any GA, both central and non-central, we conducted hierarchical training on only the referable AMD subsets (classes 3 and 4 on the 4-step scale), which consist of the most prominent AMD features. We found that the intermediate and advanced AMD cases in this subset were mostly separable within the tSNE-defined feature space, and that the intermediate AMD images that grouped with advanced AMD samples exhibited features of GA (Figures 5a-5c).

To more objectively delineate the feature pockets that define the GA phenotype, rather than human-defined demarcations between intermediate and advanced AMD classes, we performed spherical K-means clustering, which uses NPID-defined feature vectors to segregate fundus images into K classes within the 64-D feature space based on encoded vector differences (reflecting visual distinctions) rather than human labels. Using K=6 to correspond to the 6 fine-grained classes within the referrable AMD subset (classes 7-12), we found three clusters (Clusters A, B, and C) among eyes with intermediate AMD that correspond to variable degrees of GA (Figures 5a-5b), including non-central GA (Figure 5c, top row), as well as cases with central GA that should have been labeled as class 10, but were possibly mislabeled by human graders (Figure 5c, bottom row).

Because the advanced AMD hierarchical subset predictably demonstrated the greatest separability between its three fine-grained 9+3-step classes (classes 10-12), we also performed spherical K-means clustering on this subset's feature vectors using K=3 to correspond to these 3 classes (Figures 5d-5e). Here, we discovered one of the three clusters (cluster C) to contain 75% of class 10 (central GA) and class 12 (central GA + CNV) fundus images. Sampling from class 11 images within this unbiased cluster also revealed images with non-central GA, even though class 11 was agnostic to GA presence (Figures 5f). We did not locate class 11 images with obvious central GA. Thus, hierarchical learning and K-means clustering using NPID may enable unbiased, data-driven discovery of AMD phenotypes such as GA which are not specifically encoded by human-assigned AMD severity labels.

Non-AMD phenotype discovery

To identify other physiologic or pathologic phenotypes beyond AMD features, we performed K-means clustering on all training images using a K-value of 4, based on the presence of 4 coarse classes in the 4-step severity scale. We observed one cluster (Cluster A) which correspond to images with no AMD, and three other clusters (Clusters B, C, and D) which appear to straddle AMD classes, suggesting that these latter groups may be distinguished by features unrelated to AMD pathophysiology (Figures 6a-6b). A closer examination of cluster B images near the border between AMD and non-AMD classes revealed eyes with a prominent choroidal pattern known as a tessellated or tigroid fundus appearance (Figure 6c) – a feature associated with choroidal thinning and high myopia [32]. Cluster C images near this border contain fundi with a blonde appearance (Figure 6d), often found in patients with light-colored skin and eyes, or in patients with ocular or oculocutaneous albinism [33]. Images from cluster D in this area showed poorly-defined fundus appearances that were suspicious for media opacity (Figure 6e). To determine if this cluster may include eyes with greater degrees of lens opacity, we overlaid the main tSNE plot with labels for nuclear sclerosis, cortical cataracts, or posterior sub-capsular opacity from corresponding slit lamp images obtained in AREDS, and found that eyes in cluster D corresponded to a higher degree of both nuclear and cortical cataracts (Supplemental Figure 2). Hence, fundus images contain other ophthalmologically-relevant information that are not constrained to the retina, and K-means clustering of retinal images can also identify eyes with tessellated or blonde fundi as well as visually-significant cataracts.

To explore other potential phenotypes not related to AMD, we also investigated images with no AMD that are grouped with those with more advanced stages of AMD (Figure 6f). Among these images, we found examples of eyes with asteroid hyalosis – vitreous opacities consisting of calcium and lipid deposits, as well as camera artifacts such as lens flare or dirt (Figures 6g-6i), which may resemble AMD features to an nonexpert human or CNN that was not trained to identify these conditions. Our findings show that unsupervised NPID networks has the capacity for unbiased, data-driven discovery of both AMD features that were not encoded in the 4-step or 9+3 step human labels, as well as non-AMD phenotypes such as camera artifacts (lens flare or dirt), media opacity (nuclear cataracts or asteroid hyalosis), and choroidal appearance (tessellated or blonde fundus).

# DISCUSSION

In this study, we successfully trained an unsupervised neural network to grade AMD severity from fundus photographs, determined AMD features that drove network behavior, and identified novel pathologic and physiologic ocular phenotypes, all without the bias and constraints of human-assigned labels. NPID outperformed or closely matched the performance of published supervised networks and human experts in grading AMD severity on a 4-step scale (none, early, intermediate, and advanced AMD) [20], and in binary classification of advanced AMD (CNV or central GA) [14] and referable AMD (intermediate or advanced AMD) [20]. Our unsupervised network also outperformed a supervised network that was trained with both fundus images and genotype data on a custom 3-step classification of class 1, class 2-8, and class 9-12 on the 9+3-step severity scale (65% vs. 56-60)[34]. Our results suggest that even without human-generated labels, unsupervised learning can achieve predictive performance similar to expert human and supervised neural networks.

Unsupervised learning using NPID has significant advantages over supervised learning. First, eliminating the need for labor-intensive annotation of training data vastly enhances scalability and removes human error or biases. Also, NPID predictions resemble ophthalmologists more closely than do supervised networks (Figure 2d). Like humans, the unsupervised NPID network considers the AMD severity scale as a continuum and the relationship of adjacent classes. By contrast, supervised algorithms generally assume independence across classes, are susceptible to noisy or mislabeled images, and may produce more egregious misclassifications. Because the NPID algorithm groups images by visual similarity rather than class labels, inaccurate predictions can be salvaged by other nearest neighbors during group voting.

Another advantage of NPID is its versatility across different labeling schemes (2-step, 4-step, or 9+3-step), whereas distinct supervised networks must be trained or retrained for these different labeling splits, or for cross-study comparisons. Because unsupervised learning is label agnostic, the same network can be evaluated for different classification tasks, and its performance readily compared with other networks or human experts as we showed in our study.

Also, NPID predictions using locally-defined weighted k-NN voting are not dominated by overrepresented classes because local neighborhoods are populated with sufficient class homogeneity (especially with k=50 used here). This is an advantage of unsupervised over supervised approaches, as the latter trains neurons to drive overrepresented class predictions more than other classes. The training dataset used in our study exceeded the size of those used in training other supervised networks (70,349 [this study] vs 56,402 [14], 5664 [20], or 28,135 [34]). These previous studies often excluded stereoscopic duplicates or same eye images from different visits to avoid associating non-relevant fundus features such as optic disc shape or vessel patterns with any given class. Because unsupervised learning is feature-driven, and not class-driven, our network could exploit the entire available AREDS dataset which improves its coherence across different features. For instance, the testing subset used in one prior study overrepresented eyes with intermediate (33%) and late AMD (33%), and underrepresented early AMD (3%), which deviates from the skewed distribution of these classes in the full dataset (50%, 20%, 15%, and 15% across the 4-step classes, respectively), and may account for the higher reported performance of some supervised networks which are susceptible to overrepresentation bias without appropriate measures to counterbalance such as class-aware sampling or weighted loss function [35,36,13]. A similar image subset selection with a higher frequency of late AMD images than the full dataset (33% vs. 6%) could also explain the higher unbalanced accuracy and kappa for another supervised network compared to NPID, despite a lower balanced accuracy, true positive rate, and false positive rate [14].

Our findings are remarkable because while the unsupervised network was trained without labels, its performance was validated using human-assigned categories, analogous to testing students on a topic that was never taught to them. In contrast to supervised networks that were originally trained with these labels, the unsupervised network had no a priori knowledge of the classification schema, many aspects of which are defined by humans using somewhat arbitrary rationale for taxonomy that may not reflect an actual distinction in disease pathophysiology, such as distinguishing central from non-central GA due to its impact on patients' visual function and quality of life. This likely explains why NPID performed better on binary or 4-step classification of AMD severity, which more likely presents true pathophysiologic distinction, than on the finer 9+3-step AMD severity scale, where subtle differences in phenotype such as drusen size or pigmentation are arbitrarily categorized into distinct, human-defined classes. For example, our tSNE visualizations demonstrated a clear separation between no AMD and advanced AMD, but not between early-AMD classes (Figures 3a-3b), for which human grader performance is also the worst [20]. Also, the fine-grained class prediction result for each hierarchical learning setup trained on an individual 4-step class subset is consistent with the confusion matrices derived from training on the full dataset. One or two of the 9+3-step classes dominate each 4-step class due to the low visual variability among them. These findings support the need to reevaluate these finer class definitions using more unbiased, data-driven methodologies.

In our study, we probed the NPID network's behavior and found that AMD features such as drusen area drove predictions of AMD severity more than drusen size or type, or area of pigmentary changes. Using hierarchical learning and spherical K-means clustering, we also identified eyes with non-central GA among those with intermediate or advanced AMD based on proximity to eyes with central GA (class 10), even though this feature is not encoded in the human-labeled AMD severity scales. Our findings suggest that unsupervised learning can more objectively identify certain AMD phenotypes such as drusen area or GA presence which may better reflect disease pathophysiology, and enable the development of more unbiased, data-driven classification of AMD severity or subtypes that could better predict disease outcomes than human-assigned grades. Interestingly, K-means clustering also identified images with central GA that appeared mislabeled as intermediate AMD, further highlighting the ability of an unsupervised network to discover miscategorized images in ways that label-driven supervised learning cannot.

Another interesting feature of unsupervised learning is the ability to identify non-retinal phenotypes from fundus images, including camera artifacts (lens dirt or flare), media opacity (cataracts or asteroid hyalosis), and choroidal patterns (tessellated or blonde fundus). While we identified these features by spherical K-means clustering using a K-value of 4, additional cluster resolution could unveil additional pathologic or physiologic phenotypes. Future studies using von-Mises mixture models for spherical K-means clustering, which do not assume identical cluster size, may enable smaller, localized clusters of phenotypic groupings to be identified. Thus, the application of NPID may not be limited to AMD grading, and its potential supersedes that of supervised networks that are limited to the classification task for which it is trained.

Because fundus photographs exhibit very little visual and semantic variability compared to natural object images, we found that preprocessing by Laplacian-filtering transformed the spatial frequency power spectra of fundus images to better resemble natural objective images (Supplemental Figure 3), and that pretraining with ImageNet increased network performance for the 9+3-step and 4-step tasks by 100% and 33% (data not shown), respectively. This performance improvement implies that discriminating features relevant to the task were difficult to learn directly from the fundus photos, but improves with transfer learning, as seen on tSNE comparisons with and without pretraining which demonstrate extra learned features that better correlate with intermediate AMD classes (data not shown).

In summary, we trained an unsupervised network using NPID for automated grading of AMD severity from fundus photographs without the need for human-generated class labels, achieving balanced class accuracies similar to or exceeding human ophthalmologists and supervised networks that require labor-intensive manual annotations and susceptible to human error and biases. The NPID algorithm exhibits versatility across different labeling schemes without the need for retraining and is less susceptible to class imbalances, overrepresentation bias, and noisy or mislabeled images. Importantly, unsupervised learning provides unbiased, data-driven discovery of both AMD-related and other ocular phenotypes independent of human labels, which can provide insight into disease pathophysiology, and pave the way to more objective and robust classification schemes for complex, multifactorial eye diseases.

# METHODS

**AREDS Image Dataset**

The AREDS design and results have been previously reported [30]. The study protocol was approved by a data and safety monitoring committee and by the institutional review board (IRB) for each participating center, adhered to the tenets of the Declaration of Helsinki, and was conducted prior to the advent of the Health Insurance Portability and Accountability Act (HIPAA). Digitized AREDS color fundus photographs and study data were obtained from the National Eye Institute's Online Database of Genotypes and Phenotypes website (dbGaP accession phs000001, v3.p1.c2) after approval for authorized access, and exemption by the IRB.

**Data Preprocessing**

Fundus images were down-sampled to 224x224 pixels along the short edge while maintaining the aspect ratio as similarly done in past literature [23]. Fundus images were also preprocessed with a Laplacian filter applied in each of the red-green-blue (RGB) color dimensions to better emulate the properties of more natural images of everyday scenes and objects (Supplemental Figure 3). Laplacian filtering is the difference of two Gaussian-filtered versions of the original image. In this instance, it is the original fundus image (effectively, a Gaussian-filtered image with no blur) subtracted by the image Gaussian-filtered with a standard deviation (SD) of 9 pixels in each of the RGB color channels. Fundus photographs exhibit approximately the 1/f power distribution of natural images of everyday scenes and objects [37,38] but with more low-frequency than high-frequency information (Supplemental Figure 3a). The Laplacian-filtered fundus images more closely resembles that of natural statistics (Supplemental Figure 3b).

**Network Pretraining**

A CNN can transfer knowledge from one image dataset to another by using the same or similar filters [2]. Unlike natural images that contain a variety of shapes and colors that are spatially distributed throughout the image, fundus photographs are limited by shared fundus features such as the optic disc and retinal vessels, as well as the restricted colors of the retina and retinal lesions. This in turn limits the variability of the filters learned by the network. Thus, to transfer learning from a higher variety of discriminable features, we pretrained the network using the large visual database ImageNet. Pretraining affects the ideal size for the final layer feature vector in NPID, which depends on the complexity of the filters learned from the task. Our analysis revealed that an ideal size of 64 dimensions for our pretrained model maximized the performance gained from transfer learning (data not shown).

**NPID Training & Prediction**

NPID discriminates unlabeled training images using instance-based classification of feature vectors in a spherical feature space. At its core, NPID is a backbone network (ResNet50) whose logit layer is replaced with a fully connected layer of a given size (64 dimensions), and an L2-normalization function on the output feature vectors. The vectors computed for the training images are stored and compared from the previous loop of the data to determine how to update the network.

We mapped every fundus image instance $x_i$ to a representational feature vector $f_i$ within a high-dimensional hypersphere (Eqn. 1), with similar images activating similar filters mapped to nearby feature vectors, the angular distance between which are defined by their dot product (Eqn. 2).

$$f_{i_{norm}} = \frac{f_i}{|f_i|_2} \qquad (1)$$

$$d_{i,j} = f_{i_{norm}}^T f_{j_{norm}} = |f_{i_{norm}}| * |f_{i_{norm}}| * \cos(\theta_{i,j}) = \cos(\theta_{i,j}) \qquad (2)$$

The loss function for distinguishing each instance is defined by the difference between one image's feature vector and every image's stored vector from the previous loop over the data. Since it is computationally infeasible to compare against all pairs of images relative to a given image, we use 4000 negative pairs and 1 positive pair. During backpropagation, this error updates every filter in the network for better instance discrimination. Since the final layer feature vectors are all mapped to a high-dimensional unit sphere, the loss function (Eqn. 4) can minimize the distance between the current $f_{i_{norm}}$ and its corresponding stored $v_{i_{norm}}$ (i.e., 1 positive pair), while maximizing the distance between $f_{i_{norm}}$ and many other irrelevant $v_{norm}$ (i.e., 4000 negative pairs). This process is shown in Figure 1, as images are encoded with feature vectors and distributed along the spherical feature space using the loss function.

$$P(i|v_{norm}) = \frac{e^{\frac{v_{i_{norm}}^T f_{i_{norm}}}{\tau}}}{\sum_{j=i}^{n} e^{\frac{v_{j_{norm}}^T f_{i_{norm}}}{\tau}}} \quad (3)$$

$$J(\theta) = -E_d[\log P(i,v)] - m * E_n[\log(1 - P(i,v'))] \quad (4)$$

After each iteration of learning, each input image's $f_{i_{norm}}$ is pushed away from its comparison vectors before being stored in memory bank After training, predictions are made through a weighted k-Nearest Neighbors (wkNN) voting function (Eqn 5), in which votes are tallied per class c and a label is assigned based on the labels of the closest neighbors, weighted by their distance in feature space to the query's feature vector. Thus, even if a neighborhood is overrepresented by a given class, a local pocket of an underrepresented class can still dominate votes if the testing feature vector falls close to that pocket.

$$w_c = \sum_{i \in N_k} e^{\frac{v_{norm}^T f_{i_{norm}}}{\tau}} * (c_i = c) \qquad (5)$$

**Measurement of Network Performance**

We evaluated trained network performance by measuring the overall testing accuracy on a novel group of images across both the 2-Step classification and 4-Step AMD severity scale. We chose the epoch that yielded the best validation accuracy using the weighted k-nearest neighbor (k-NN) classification voting scheme from Equation 5. This included the unbalanced and balanced accuracy of the best epoch, and metrics such as Cohen's kappa, true positive rate, and false positive rate are also reported. Unbalanced accuracy is the average accuracy across all samples, whereas balanced accuracy is the average class accuracy [39,40]. While both accuracy metrics are relevant and positively highlight the performance of NPID, balanced accuracy is less biased to skewed class distributions by weighting underrepresented class scores as equally as overrepresented ones, and is more appropriate for comparing performance across different subsets of the same data as in our study.

**t-SNE visualization & Search Similarity**

To assess neighborhoods of learned features, we evaluated search similarity and t-Distributed Stochastic Neighbor Embedding (t-SNE) visualizations. Search similarities show how a given query image's severity is predicted based on nearest neighbor references, and t-SNE visualizations show us how all the fundus images are distributed across neighborhoods of visual features chosen by the network. Specifically, tSNE maps feature vectors from high-dimensional to low-dimensional coordinates while approximately preserving local topology. Here, we map the encoded 64D features onto 2D coordinates, wherein coordinates that are near each other in 2D are also near each other in the original feature space, meaning they are similarly encoded because they share visual features. Thus, we can color each 2D coordinate by the known labels for each fundus image in the training set to observe which images are encoded near to each other and what visual groupings emerge from these locally similar encodings. This process is label agnostic, so evaluation across multiple domains of labels (e.g. 2-step AMD severity, 4-step AMD severity, drusen count, media opacity, etc.) is possible without retraining, unlike a supervised network.

**Hierarchical Learning**

Because NPID appears more suitable for coarse-level than fine-level classification across dependent classes, we split up the 9+3-step dataset into each of the 4-step classes. We trained the NPID network on only no, early, intermediate, or advanced AMD images, then evaluated NPID's ability to discriminate between the three fine 9+3-step classes within each coarse 4-step class to identify which of the 9+3-step classes appear to show less visual discriminability than the grading rubric suggests.

**Spherical K-Means Clustering**

To locate specific, notable training images aside from exhaustive similarity searches of random query images, we employed spherical K-means to identify clusters of training images of interest. For conventional K-means clustering, the algorithm groups feature vectors into k distinct equally-sized gaussian-distributed groups based on the distances of the feature vectors to the approximated group centers [41,42]. Spherical K-means differs by calculating the distance along a sphere, instead of directly through Euclidean space, which is more suitable for NPID because it maps images into vectors on a sphere [43]. Mapping of K-means-defined labels onto the pre-existing tSNE helps to identify regions that are notably defined by or distinct from the original labels for further analysis.

# Figure Legends

**Figure 1: Schematic of NPID training & testing.**
Schematic diagram of the process by which Non-Parametric Instance Discrimination (NPID) trains an unsupervised neural network to map preprocessed fundus images to embedded feature vectors. The feature vectors and associated AMD labels are used as a reference for queried severity discovery through neighborhood similarity matching. The NPID network can then be analyzed to measure balanced accuracy in AMD severity grading, explore visual features that drive network behavior, and discover novel AMD-related features and other ocular phenotypes in an unbiased, data-driven manner.

**Figure 2: Comparison of unsupervised NPID performance with supervised networks and human experts.**
**(a-c)** Comparisons of the unsupervised NPID network performance with supervised networks and human ophthalmologists published by *Peng et al. [14] and #Burlina et al. [20] for binary classification of advanced AMD (a) or referable AMD (b), as well as the 4-step AMD severity scale (c). **(d)** Comparison of confusion matrices of our unsupervised NPID network with the supervised network and human expert gradings reported in #Burlina et al. [20] for the 4-step AMD severity scale task. **(e)** Confusion matrix of the NPID network on the 9+3-step AMD severity classification task.

**Figure 3. Unsupervised NPID clusters fundus images based on visual similarity**
t-Distributed Stochastic Neighbor Embedding (t-SNE) visualizations of NPID feature vectors colored by **(a)** 4-step and **(b)** 9+3-step AMD severity labels, where each colored spot represents a single fundus image with AMD severity class as described in the legend and Supplemental Figure 1. **(c)** Representative search similarity images for successful and failed cases for the 9+3-step AMD severity scale task. The leftmost column corresponds to the query fundus image, while the next 5 images on each row correspond to the top 5 neighbors as defined by network features. The colored borders and numeric labels for each image define the true class label defined by the reading center for AREDS, and correspond to the color scheme in Supplemental Figure 1.

**Figure 4. AMD-related fundus features that drive unsupervised NPID network predictions.**

t-Distributed Stochastic Neighbor Embedding (t-SNE) visualizations of NPID feature vectors colored by AREDS reading center labels for AMD-related fundus features, with corresponding stacked bar plots showing ratio of each label across the 4-step AMD severity classes. Labels include **(a)** drusen area, **(b)** maximum drusen size, **(c)** reticular drusen presence, **(d)** soft drusen type, **(e)** hyperpigmentation area, **(f)** depigmentation area, **(g)** total geographic atrophy (GA) area, and **(h)** central GA area. Category definitions for each fundus feature are shown in Supplemental Table 1.

**Figure 5. Data-driven discovery of central and non-central geographic atrophy.**
t-Distributed Stochastic Neighbor Embedding (t-SNE) visualizations of NPID feature vectors colored by **(a)** 9+3-step AMD severity labels and **(b)** spherical K-means cluster labels with K=6, based on hierarchical learning using only fundus images with referable AMD (intermediate or advanced AMD). A selection (outlined area) of intermediate AMD cases (classes 7-9) adjacent to advanced AMD cases (classes 10-12) from clusters A-C show **(c)** fundus images with non-central GA (top row) and central GA (bottom row). t-SNE visualizations of NPID feature vectors colored by **(d)** with 9+3-step AMD severity labels and **(e)** spherical K-means cluster labels with K=3, based on hierarchical learning using only fundus images with advanced AMD (classes 10-12). A selection (outlined area) of CNV cases (class 11) adjacent to images with central GA with or without CNV (classes 10 and 12) from cluster C show **(f)** non-central GA.

**Figure 6. Data-driven discovery of ophthalmic features.**
t-Distributed Stochastic Neighbor Embedding (t-SNE) visualizations of NPID feature vectors colorerd by **(a)** 4-step AMD severity labels and **(b)** spherical k-means (K=4) cluster labels. Fundus images that straddle no AMD vs. early, intermediate, or advanced AMD within K-means cluster B (yellow-purple circle), cluster C (teal-blue circle), and cluster D (green-red circle), corresponded to fundus images with **(c)** tessellated fundus, **(d)** blonde fundus, and **(e)** media opacity. **(f)** t-SNE visualization of 9+3-step AMD severity labels with a selection (outlined areas) of fundus images with no AMD (class 1) located within clusters of early, intermediate, or late AMD classes corresponded to fundus images with **(g)** asteroid hyalosis, **(h)** camera lens flare, and **(i)** camera lens dirt.

**Supplemental Figure 1. 9+3-step and 4-step AMD severity scales & data distribution**.
**(a)** Dendogram showing representative images from each of the 9+3-step AMD severity classes as defined by the reading center for AREDS, and the simplified 4-step AMD severity classes including no AMD (blue), early AMD (aqua), intermediate AMD (yellow), and advanced AMD (red). **(b-c)** Histogram plots across training labels for the (c) 9+3-step and (c) 4-step AMD severity scales.

**Supplemental Figure 2. Cluster breakdowns for lens phenotypes.**
**(a-c)** t-Distributed Stochastic Neighbor Embedding (t-SNE) visualizations of NPID feature vectors colored by AREDS reading center labels for (a) nuclear cataract severity, (b) cortical cataract severity, and (c) posterior subcapsular cataract severity. Below each t-SNE plot are stacked bar plots showing histogram distribution across the spherical k-means (k=4) clusters from Figure 6b. Category definitions for cataract severities are shown in Supplemental Table 1.

**Supplemental Figure 3. Preprocessing steps for difference with Gaussian filtering.**
**(a)** Comparison of preprocessing steps on representative fundus image, and **(b)** corresponding azimuthally-defined 1D power spectrum. Blue, orange, and green power spectra lines correspond to the images in (a), while the red line corresponds to the power spectrum of natural images.

**Supplemental Table 1. Class definitions for AREDS Reading Center labels.** Descriptive tables detailing label definitions for (a) drusen area, (b) max drusen size, (c) reticular drusen presence, (d) soft drusen type, (e) hyperpigmentation area, (f) depigmentation area, (g) total geographic atrophy (GA) area, (h) central GA area, (i) nuclear cataracts severity, (j) cortical cataracts severity, (k) posterior sub-capsular cataracts severity. In the table definitions, C, I, and O correspond to groups of open circles, where C=Central, I=Inner, and O=Outer. Their numbers correspond to the Disc Diameter (DD) in relation to the average Disc Area (DA), C0=0.042 DD, C1=0.083 DD, C2=0.167 DD, I1=0.120 DD, I2=0.241 DD, O1=0.219 DD, O2=0.439 DD.